\documentclass{amsart}
\usepackage{amssymb}
\usepackage{color}

\newtheorem{theorem}{Theorem}

\newtheorem*{theorem*}{Theorem}
\newtheorem*{lemma*}{Lemma}

\theoremstyle{definition}
\newtheorem{definition}{Definition}
\newtheorem{remark}{Remark}

\newtheorem*{remark*}{Remark}
\newtheorem*{example*}{Example}
\DeclareMathOperator*{\argmin}{arg\,min}
\theoremstyle{plain}

\begin{document}
  \title[
Universality of almost periodic orbits
]{
Universality of almost periodic orbits in 
certain composite functions
}

\author{Chikara Nakayama} 
\address{Graduate School of Economics, Hitotsubashi University, 2-1 Naka, Kunitachi, Tokyo 186-8601, Japan} 
\email{c.nakayama@r.hit-u.ac.jp}

\author{Tsuyoshi Yoneda} 
\address{Graduate School of Economics, Hitotsubashi University, 2-1 Naka, Kunitachi, Tokyo 186-8601, Japan} 
\email{t.yoneda@r.hit-u.ac.jp}




\begin{abstract} 
We consider composite functions in the elementary algebraic framework.
Without any use of the Fourier transform,
 we find almost periodic orbits which suitably characterizes 
certain
composite functions.
In particular, 
we provide special composite functions that 
tend asymptotically to almost periodic orbits. 
\end{abstract}

  \maketitle

\section{
Introduction
}

In this paper we consider universality of almost periodic orbits from elementary algebraic point of view.
More precisely,
without any use of the Fourier transform,
 we show that
certain composite functions
can be characterized by 
almost periodic orbits.
Furthermore, we provide  
special composite functions
that tend asymptotically to almost periodic orbits. 
Let us formulate them more precisely.
Let $d\in\mathbb{Z}_{\geq 1}$ called a dimension. 
For a map
\begin{equation}\label{Phi}
    \Phi:[-1,1]^d\to[-1,1]^d,
\end{equation}
we generate a 
composite function $y:\mathbb{Z}_{\ge 0}\to[-1,1]^d$ as follows:
\begin{equation*}
y(t+1)=\Phi(y(t))
\quad\text{for an initial data}\quad y(0)\in[-1,1]^d.
\end{equation*}
Let $\gamma\in\mathbb{R}_{> 1}$ be the corresponding Lipschitz constant, and assume it is finite, as follows: 
\begin{equation*}
\gamma:=\sup_{\stackrel{W,W'\in[-1,1]^d,}{W\not=W'}}\frac{|\Phi(W)-\Phi(W')|}{|W-W'|}<\infty.
\end{equation*}
Note that, for $\Phi^t:=\underbrace{\Phi\circ\Phi\circ\cdots\circ\Phi}_{t\ \text{times}}$,  
\begin{equation}\label{optimal estimate}
    |\Phi^t(W)-\Phi^t(W')|\leq \gamma^t|W-W'|
    \quad\text{for}\quad W,W'\in [-1,1]^d.
\end{equation}
Then the main theorems are as follows:
\begin{theorem}\label{Theorem 1}
There exist 
approximating functions $y^*_K:\mathbb Z_{\ge 0} \to [-1,1]^d$\\ ($K \in \mathbb{Z}_{\ge 1}$), 
constants: $\{L_K\}_{K\in \mathbb{Z}_{\ge 1}}\subset \mathbb{Z}_{\geq 1}$, $\{T_K\}_{K\in \mathbb{Z}_{\ge 1}}\subset \mathbb{Z}_{\geq 1}$ and\\
$\{M_K\}_{K \in\mathbb{Z}_{\geq1}}\subset \mathbb{Z}_{\geq 0}$ 
such that 
\begin{equation*}
y^*_K(t)=\sum_{m=0}^{M_K} a_m\sin\left(\frac{2\pi mt}{L_K}\right)+b_m\cos\left(\frac{2\pi mt}{L_K}\right)
\quad (\{a_m\}_m, \{b_m\}_m\subset\mathbb{R}^d)
\end{equation*} 
for $t\geq T_K$ and 
\begin{equation}
\label{ineq}
\displaystyle|y^*_K(t)-y(t)|\leq C(t)\gamma^t
\frac{\sqrt d}{K}\quad \text{for}\quad t\geq 0,
\end{equation}
where $C(t)=\displaystyle\frac{2(\gamma-\gamma^{-t+1})}{\gamma-1}+\gamma^{-t}$.
Since $\displaystyle C(t)< \frac{2\gamma}{\gamma-1}$,
this estimate \eqref{ineq} is only on the constant worse (c.f. \eqref{optimal estimate}). Thus, $y^*_K$ seems almost optimal for approximating $y$.
\end{theorem}
Next we provide a conditional characterization of $y$.
Before that, we need to give a definition of  almost periodic functions \cite[(1.2)]{C}.
\begin{definition}\label{def of ap}
  A function $f:\mathbb{R}\to\mathbb{R}$ is called almost periodic function, if for any $\epsilon>0$,
there exist $M>0$, $\{\lambda_k\}_{k=1}^M\subset\mathbb{R}$ and $\{c_k\}_{k=1}^M\in\mathbb{C}$
such that 
\begin{equation*}
    \sup_{x\in\mathbb{R}}\left|f(x)-\sum_{k=1}^{M} c_ke^{i\lambda_k x}\right|<\epsilon.
\end{equation*}
\end{definition}
Note that, due to \cite[Theorem 1.20]{C}, any almost periodic function $f$ can be expressed as 
\begin{equation*}
    f(x)=\sum_{k=1}^\infty c_ke^{i\lambda_k x}
\end{equation*}
with uniquely determined $\{c_k\}_k\subset\mathbb{C}$
and $\{\lambda_k\}_k\subset\mathbb{R}$.
\begin{theorem}\label{Theorem 2}
Let $\mathcal L_{K,K'}$ be the least common multiple of periods $L_K$ and $L_{K'}$.
For $\{K_j\}_j$ ($K_1<K_2<\cdots<K_j<\cdots$),
we re-select $T_{K_1}, T_{K_2},\cdots, T_{K_j},\cdots$ such that
\begin{equation*}
T_{K_1}\leq T_{K_2}\leq \cdots\leq T_{K_j}\leq \cdots\quad\text{and}\quad 
T_{K_{j+1}}-T_{K_j}\ \text{is divisible by}\ L_{K_j}.
\end{equation*}
 If 
\begin{equation}\label{converging condition}
\sum_j(2T_{K_{j+1}}+2\mathcal L_{K_{j+1},K_j}+1)\frac{\gamma^{T_{K_{j+1}}+\mathcal L_{K_{j+1},K_j}}}{K_j}<^\exists\!\! C^*<\infty, 
\end{equation}
then there exists an almost periodic function $ap(t)$
such that the following convergence holds:
\begin{equation}\label{conv}
\sup_{0\leq t\leq \mathcal L_{K_{j+1},K_j}}|y(t+T_{K_j})-ap(t)|\to 0\quad (j\to\infty).
\end{equation}
\end{theorem}
Finally, we provide special composite functions that 
tend asymptotically to almost periodic orbits.
Let 
$\Phi:[-1,1]^d\to[-1,1]^d; (w_1,w_2,\cdots,w_d)^T\mapsto (x_1,x_2,\cdots,x_d)^T$
be such that
\begin{equation}\label{delay coordinate}
w_1=x_2,\quad w_2=x_3,\cdots, w_{d-1}=x_d.
\end{equation}
In this case we can rephrase $y:\mathbb{Z}_{\geq 0}\to[-1,1]^d$ to $z:\mathbb{Z}_{\geq -d+1}\to[-1,1]$ as follows: 
\begin{equation*}
    y(t)=(z(t),z(t-1),\cdots,z(t-d+1))^T.
\end{equation*}
To state the third theorem, we use the following autoregressive model (see \cite{CFY} and more generalized versions, see \cite{CGC,E,OO} for example):
\begin{equation}\label{AR}
z(t)=\sum_{\ell=1}^dp_{\ell}z(t-\ell)
\end{equation}
for 
$\{p_\ell\}_{\ell=1}^d\subset\mathbb{R}$. 
\begin{theorem}\label{AR model}
If $\Phi:[-1,1]\to[-1,1]$ generates an autoregressive model \eqref{AR}, then  there is a unique almost periodic function $ap$ such that the corresponding $y$ satisfies
$y(t)\to ap(t)$ $(t\to\infty)$. 

\end{theorem}
\begin{remark}
It is
an open question whether this $ap$ coincides with $ap$ that is in \eqref{conv} (or, before that, whether the condition \eqref{converging condition} holds).
\end{remark}

\section{Proof of theorems}

\noindent
{\it Proof of Theorem \ref{Theorem 1}.}
\
First let us discretize the range $[-1,1]$ as follows:
For $K\in\mathbb{Z}_{\geq 1}$,  we choose $A:=\{a_k^K\}_{k=0}^{K}\subset[-1,1]$ such that 
\begin{equation*}
    a_k^K:=\frac{2k}{K}-1.
\end{equation*}

Then we have 
\begin{itemize}
\item
$-1=a_{0}^K<a_{1}^K<a_{2}^K<\cdots
<a_{K}^K=1$,
\item

$|a_{k-1}^K-a_{k}^K|
=\displaystyle\frac{2}{K}$.

\end{itemize}

We now discretize $y$ as follows:
\begin{equation}\label{discrete}
\bar y_K(t):=\argmin_{a\in A^d}|y(t)-(a-0)|,
\end{equation}
where $a-0:=a-\varepsilon$ for any sufficiently small $\varepsilon>0$.
Note that 
\begin{equation*}
    |y(t)-\bar y_K(t)|\leq \frac{\sqrt d}{K}.
\end{equation*}
We classify patterns in $\bar y$.   
Let $\sigma_n^K$ ($n=1,2,\cdots,N$) be a permutation operator, namely, a map 
\begin{equation*}
\sigma_n^K:\{1,2,\cdots, d\}\to\{-1,a_{1}^K,a_{2}^K,\cdots,a_{K-1}^K,1\}
\end{equation*}
with $\sigma_n^K\not=\sigma_{n'}^K$ ($n\not=n'$), 
and we impose the following two conditions for determining $N$:
\begin{equation}\label{all patterns}
\begin{cases}
\text{For any $t\in\mathbb{Z}_{\ge 0}$, there is $n\in\{1,\cdots,N\}$ such that }
\text{$\sigma_n^K=\bar y_K(t)$},\\
\text{For any  $n\in\{1,\cdots,N\}$ there is $t\in\mathbb{Z}_{\ge 0}$ such that }
\text{$\sigma_n^K=\bar y_K(t)$.} 
\end{cases}
\end{equation}
Note that  $N\leq (K+1)^d$ due to the sequence with repetition.
We define a next value $\sigma_{k(n)}^K$ of $\sigma_n^K$ as follows:
For any $K\in\mathbb{Z}_{\geq 1}$ and $n\in\{1,2,\cdots, N\}$,  we choose a $k(n)\in\{0, 1,2,\cdots, K\}$ such that 
\begin{equation}\label{next value}
\sigma_{k(n)}^K=\bar y_K(t+1)\quad\text{and}\quad \sigma_n^K=\bar y_K(t)\quad\text{for some}\quad t\in\mathbb{Z}_{\ge 0}.
\end{equation}
Now we construct approximating functions 
$\{y^*_K(t)\}_{t\geq 0}$ inductively
from initial  data $\bar y_K(0)$.
First, let 
\begin{equation*}
y^*_K(0)=\bar y_K(0).
\end{equation*}
Assume that for $y^*_K(t)$ $(t \ge0)$ is already defined and that  
\begin{equation}\label{induction condition}
y^*_K(t)=\sigma^K_n
\end{equation}
for some $n$.  Define $y^*_K(t+1)=\sigma^K_{k(n)}$ and the condition \eqref{induction condition} holds also for the $t+1$ case. 
Therefore the induction argument goes through, and then we obtain the desired sequence 
$\{y^*_K(t)\}_{t=0}^\infty$.
Next we show \eqref{ineq}.
By the right above argument, for any $t\in\mathbb{Z}_{\geq 0}$, there exists $t'\in\mathbb{Z}_{\geq 0}$ such that
\begin{equation}\label{uniform estimates}
\begin{cases}
\displaystyle
\overline y_K(t')=y^*_K(t),\quad |y(t')- \overline y_K(t')|\leq \frac{\sqrt d}{K},\\
y(t')=\Phi(y(t'-1)),\\
\displaystyle
\overline y_K(t'-1)=y^*_K(t-1),\quad
|y(t'-1)-\bar y_K(t'-1)|\leq \frac{\sqrt d}{K}.
\end{cases}
\end{equation}
Then we have 
\begin{equation*}
\begin{split}
|y^*_K(t)-y(t)|\leq
&
|\bar y_K(t')-y(t)|\\
\leq &
|\bar y_K(t')-y(t')|+|y(t')-y(t)|\\
\leq 
&
 |y(t')-y(t)|+\frac{\sqrt d}{K}.\\
\end{split}
\end{equation*}
Since 
$\Phi(y(t-1))=y(t)$ and \eqref{optimal estimate},
we see that
\begin{equation*}
    \begin{split}
\leq
&
\gamma |y(t'-1)-y(t-1)|+\frac{\sqrt d}{K}\\
\leq 
&\gamma (|y(t'-1)-\bar y_K(t'-1)|+|\bar y_K(t'-1)-y(t-1)|)
+
\frac{\sqrt d}{K}\\
\leq
&\gamma |y^*_K(t-1)-y(t-1)|+\gamma\frac{\sqrt d}{K}
+
\frac{\sqrt d}{K}.\\
\end{split}
\end{equation*}
Thus we have the following recurrence inequality:
\begin{equation*}
    |y^*_K(t)-y(t)|\leq \gamma|y^*_K(t-1)-y(t-1)|+\gamma\frac{\sqrt d}{K}+\frac{\sqrt d}{K},
\end{equation*}
and by the induction argument, we have 
\begin{equation*}
\begin{split}
 |y^*_K(t)-y(t)|
 &\leq 
 (2(\gamma^t+\gamma^{t-1}+\cdots +\gamma)+1)\frac{\sqrt d}{K}\\
 &=\left(\frac{2(\gamma-\gamma^{-t+1})}{\gamma-1}+\gamma^{-t}\right)\gamma^t\frac{\sqrt d}{K}.
 \end{split}
 \end{equation*}
Therefore we have  \eqref{ineq}.
Now we show that  $y^*_K(t)$ is a discretized periodic function on $t\geq T_K$ for some $T_K$. 
By the construction of $y^*_K$, 
for each $K$, we have a chain of the permutation operators as follows:
\begin{equation*}
\underbrace{
\sigma_{j}^K\rightarrow
\sigma_{k(j)}^K\rightarrow\sigma_{k(k(j))}^K\rightarrow\cdots\rightarrow
\sigma^K_{k^M(j)}}_{M\ chains}
\rightarrow\cdots
.
\end{equation*}
Since the number of permutation operators $\{\sigma_n^K\}_{n=1}^N$ is finite, 
then there exist
 $m, M\in\mathbb{Z}_{\geq 1}$ ($m<M$) such that $\sigma_{k^m(j)}^K=\sigma_{k^M(j)}^K$.
This means that there exists $T_K$ such that $y^*_K(t)$ is a periodic chain for $t\geq T_K$. We set its period as $L_K$.
This is the desired result.

\vspace{0.5cm}

\noindent
{\it Proof of Theorem \ref{Theorem 2}.}
\
We have the following estimate for the periodic functions $y^*_{K_j}(t)$ ($t\geq T_{K_j}$) and $y^*_{K_{j+1}}(t)$ ($t\geq T_{K_{j+1}}$):
\begin{equation*}
\begin{split}
&\sup_{t\geq 0}|y^*_{K_{j+1}}(t+T_{K_{j+1}})-y^*_{K_j}(t+T_{K_j})|\\
= &
\sup_{t\geq 0}|y^*_{K_{j+1}}(t+T_{K_{j+1}})-y^*_{K_j}(t+T_{K_{j+1}})|\\
=&
\sup_{0\leq t\leq \mathcal{L}_{{K_{j+1},K_j}}}|y^*_{K_{j+1}}(t+T_{K_{j+1}})-y^*_{K_j}(t+T_{K_{j+1}})|.\\
\end{split}
\end{equation*}
We have used 
divisible by $L_{K_j}$ property for the first to second,
common multiple of periods $L_K$ and $L_{K'}$ property 
for the second to last.
By applying \eqref{ineq}, we have 
\begin{equation*}
    \begin{split}
&\leq 
\sup_{0\leq t\leq \mathcal{L}_{{K_{j+1},K_j}}}|y^*_{K_{j+1}}(t+T_{K_{j+1}})-y(t+T_{K_{j+1}})|\\
&\qquad+\sup_{0\leq t\leq \mathcal {L}_{K_{j+1},K_j}}|y^*_{K_j}(t+T_{K_{j+1}})-y(t+T_{K_{j+1}})|\\
&\leq 
(2T_{K_{j+1}}+2\mathcal L_{K_{j+1},K_j}+1){\gamma}^{T_{K_{j+1}}+\mathcal L_{K_{j+1},K_j}}\frac{\sqrt d}{K_{j+1}}\\
&\qquad
+
(2T_{K_{j+1}}+2\mathcal L_{K_{j+1},K_j}+1)\gamma^{T_{K_{j+1}}+\mathcal L_{K_{j+1},K_j}}\frac{\sqrt d}{K_j}\\
& \leq
2(2T_{K_{j+1}}+2\mathcal L_{K_{j+1},K_j}+1)\gamma^{T_{K_{j+1}}+\mathcal L_{K_{j+1},K_j}}\frac{\sqrt d}{K_j}.
\\
\end{split}
\end{equation*}
By \eqref{converging condition},
 $\{y^*_{K_j}(\cdot +T_{K_j})\}_j$ is uniform convergence on $\mathbb{Z}$,
and by  Definition \ref{def of ap}, 
there is an almost periodic function $ap$ such that
\begin{equation*}\label{ap}
\displaystyle\sup_{t\geq 0}
|y^*_{K_j}(t+T_{K_j})-ap(t)|\to 0\quad(j\to\infty).
\end{equation*} 
Combining the following convergence (due to \eqref{ineq} again): 
\begin{equation*}
\begin{split}
&\sup_{0\leq t\leq\mathcal L_{K_{j+1},K_j}}|y(t+T_{K_j})-y^*_{K_j}(t+T_{K_j})|\\
= 
&\sup_{T_{K_j}\leq t\leq T_{K_j}+\mathcal L_{K_{j+1},K_j}}|y(t)-y^*_{K_j}(t)|\to 0\quad 
(j\to\infty), 
\end{split}
\end{equation*}
we have the desired \eqref{conv}.

\vspace{0.5cm}

\noindent
{\it Proof of Theorem \ref{AR model}}\quad
First we figure out the corresponding characteristic equation, that is, 
we plug the following:
\begin{equation*}
 z(t-\ell)=\mu^{d-\ell},\quad  (\mu\in\mathbb{R},\ \ell=0,1,\cdots, d)
\end{equation*}
 into \eqref{AR}.
After factorization of the $d$-th degree polynomial, then, we see that $\{p_\ell\}_\ell$ satisfies the following equality:
\begin{equation}\label{factorization}
\prod_{j=1}^d(\mu-\mu_j)
=\mu^d-\sum_{\ell=1}^{d}p_\ell\mu^{d-\ell}=0
\end{equation}
for some $\{\mu_j\}_j\subset\mathbb{C}$.
From this factorization, we have the following representation:
\begin{equation}\label{representation}
    z(t)=\sum_{j=1}^da_jt^{k_j}\mu_j^t\quad (\{a_j\}_j\subset\mathbb{C})
\end{equation}
for $t>0$, where $\{k_j\}_j\subset\mathbb{Z}_{\geq 0}$ originate from Jordan normal forms. Note that the coefficients $\{a_j\}_j$ can be determined by the initial data:
\begin{equation*}
    y(0)=(z(0),z(-1),\cdots, z(-d+1))^T.
\end{equation*}
To be more precise, we first generate
\begin{equation*}
    y(d-1)=(z(d-1),z(d-2),\cdots, z(0))^T,
\end{equation*}
by using \eqref{AR} inductively.
We plug it into \eqref{representation}
and then we obtain the simultaneous equations of the coefficients $\{a_j\}_j$.

We now separate $z$ into the almost periodic part and the decaying part.
If there exists $j$ such that $|\mu_j|>1$, or 
$|\mu_j|=1$ and $k_j\geq 1$, then $z(t)$ $(t=1,2,\cdots)$
is unbounded. Since the growing case is already excluded
due to the definition \eqref{Phi},
 the pair of $(\mu_j,k_j)$ must satisfy
\begin{equation*}
    |\mu_j|=1\quad\text{and}\quad k_j=0,\quad\text{or}\quad |\mu_j|<1,\quad\text{for every}\quad j=1,2,\cdots, d. 
\end{equation*}
Thus there exist an almost periodic function $ap(t)$
and a decaying function $R(t)$ ($R(t)\to 0$ as $t\to\infty$) such that 
\begin{equation*}
    z(t)=ap(t)+R(t).
\end{equation*}
This is the desired explicit representation.

\begin{remark}
Under the setting of \eqref{delay coordinate},
the period of $y_K^*$ should be studied more. Theoretically, we see that it can be any value from 1 to $(K+1)^d$ by a result in \cite{IMY}, which says that for any $1\le p\le (K+1)^d$, there exists a $\Phi\colon \{0,1,2,\ldots,K\}\to \{0,1,2,\ldots,K\}$ satisfying \eqref{delay coordinate} that generates a time series of period $p$.  However, in general, we observe that the period of $y^*_K$ is much shorter than $(K+1)^d$.  We can ask what the average value is of the periods for some model or how it depends on $K$.
\end{remark}

\vspace{0.5cm}
\noindent
{\bf Acknowledgments.}\  
The first author thanks J.\ C.\ for leading him to join this work.
Research of  TY  was partly supported by the JSPS Grants-in-Aid for Scientific
Research  24H00186.
Research of  CN was partly supported by the JSPS Grants-in-Aid for Scientific
Research 21K03199.



\end{document}